# Radial Based Analysis of GRNN in Non-Textured Image Inpainting

Karthik R, Anvita Dwivedi, Haripriya M, Bharath K P, Rajesh Kumar M, *Senior IEEE member*

*School of Electronics Engineering*
*VIT University*
Vellore, India 632014
tkgravikarthik@gmail.com, anvitadwivedi95@gmail.com, haripriyabandu@gmail.com, bharathkp25@gmail.com, mrajeshkumar@vit.ac.in

*Abstract---Image inpainting algorithms are used to restore some damaged or missing information region of an image based on the surrounding information. The method proposed in this paper applies the radial based analysis of image inpainting on GRNN. The damaged areas are first isolated from rest of the areas and then arranged by their size and then inpainted using GRNN. The training of the neural network is done using different radii to achieve a better outcome. A comparative analysis is done for different regression-based algorithms. The overall results are compared with the results achieved by the other algorithms as LS-SVM with reference to the PSNR value.*

*Keywords—GRNN;LS-SVM;PSNR;MSE;Radial-based analysis;Regression.*

## I. INTRODUCTION

One of the remarkable problems in image processing is to reconstruct damaged or incomplete images as much as possible. Image inpainting can also be used as a powerful technique for restoring old and scratched pictures or artworks. In the digital world, the problem of image inpainting lies at the intersection of computer graphics, image, and signal processing. Various inpainting methods have been presented like Partial Differential Equation (PDE), Total Variation (TV), Curvature Driven Diffusions (CDD), texture synthesis, exemplar-based method, Gaussian Process, Support Vector Machine (SVM), Neural network based algorithms as General Regression Neural Network(GRNN) etc...

The problem of image inpainting has been dealt as a part of a range of research work and the main question to deal is to find the way that restores the image almost perfectly. The main applications of image inpainting are in object removal and region filling [1], video restoration [2], text removal [3], digital zooming [4] and watermark removal [5]. Many inpainting methods have been introduced which include inpainting using Partial differential equations 6] which work for the diffusion-based algorithms and hence performs better for small-scale damage reduction, exemplar-based image inpainting [1], regression-based neural networks as GRNN [7] which produce good result for large-scale scratches, interpolation techniques [8] which are good for fast and accurate reconstruction of damages which are scattered in the digital image. In recent research, some unconventional ways have been tried to inpaint the digital images using Artificial Neural Networks which work as a powerful tool for black-box Modelling and are designed from real biological neural networks are used to inpaint the images. ANN's as Convolution Neural Networks (CNN) [9] and Deep Neural Networks. Dictionary-based learning with the sparse representation algorithms as K-SVD [10]. Sparse dictionary-based model are also used for adaptive patch learning[11]which can be used to inpaint images which have very large block of image missing and have to inpaint face features which cannot be done by only using the surrounding info of the missing region, the inpainting of blocks of face features comes under image completion[12].The accepted graphic programs as Adobe Photoshop has also been used using its content-aware fill feature in CS5 version [13] which can carry out image inpainting with high efficiency and low latency but is subjected to software crashes. This paper mainly focuses on radial analysis of GRNN. Regression-based methods which are selected here are based on their memorizing ability like GRNN [7], LSSVM [14].

Alilou [7] mainly focuses on training of GRNN using single radius, however using different radii for training the GRNN, we can find strongly correlating pixels for filling the missing region to yield more appropriate result. The variation in result is studied after taking radius as a parameter for neural network training to achieve optimization while real implementation. To make the future research more comfortable we made a comparative study on LSSVM and GRNN.

In section II, the methodology for radial analysis has been discussed .In section III explains two major regression models for image inpainting and Database with masks. In section VI comparison of results. Section V, concludes the results obtained from result and overall papers conclusion has been made

## II. PROPOSED METHOD

The proposed work has been divided into three algorithms. Algorithm 1, explains the method for identifying training pixels and has the flexibility of changing radius for analysis purpose. In algorithm 2, the training is done by identifying damaged and training parts, and selecting regression model. Algorithm 3, explains the final inpainting process in which output from the trained model is fitted into damaged coordinates.

Algorithm 1:

Step 1.Identify damaged part of Damaged image manually

Step 2.Create a mask matrix which denotes 1 for non damaged parts and 0 for damaged parts

Step 3.Element-wise multiplication of damaged image with mask (mask and damaged image should be in same size)

Step 4.Determine the radius (R) which can be varied from 3 to any distance less then the distance measured from nearest edge or side

Step 5.Apply R-by-R average filter for both the images.

Step 6  Subtract masked image from the original image and obtain the non zero part .

Step 7 Use the non-zero part of subtracted image for training the regression model.

The blurring of the original and masked image is done using averaging filter .The blur impacts pixels up to a certain radius. Hence the subtraction of the original blurred image with the masked blurred image is effected because of the influence of mask on the neighboring pixels .The radius of affected neighboring will be determined by kernel size in avg. filter and it's denoted by 'R'. Since the filter kernel cannot be determined less than 3 and radius is not valid outside the image hence the radius can be set in the interval of 3 to distance measured from nearest edge or side of the image.

Algorithm 2:

Step 1.The Pixels are taken from the coordinates which are obtained from Algorithm 1.

Step 2.Selecting an apt. regression model

Step 3.Train the regression model by giving (x, y) coordinates as inputs and pixels values (I[x, y]) as the target.

Step 4.Trained model is saved for inpainting purpose.

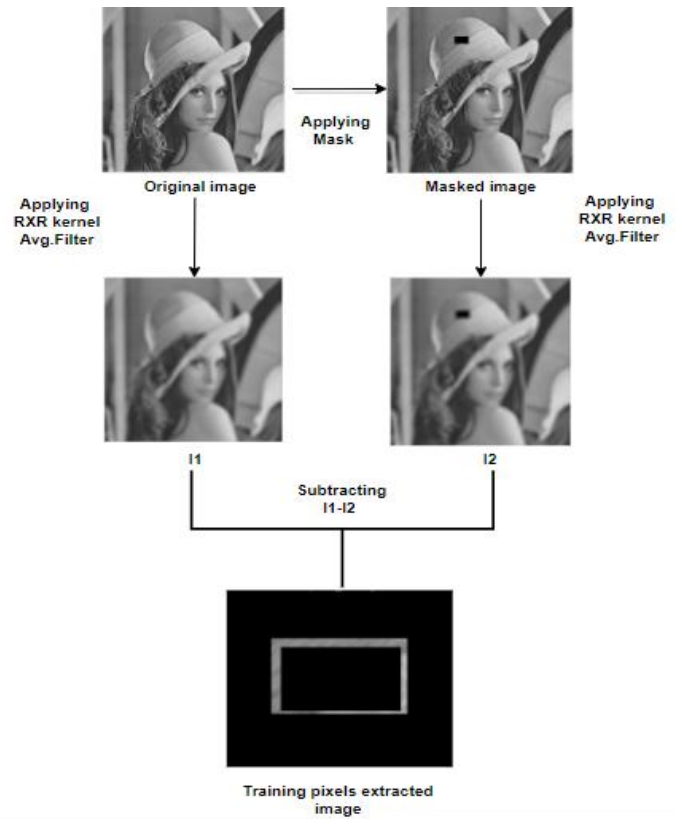

Figure.1

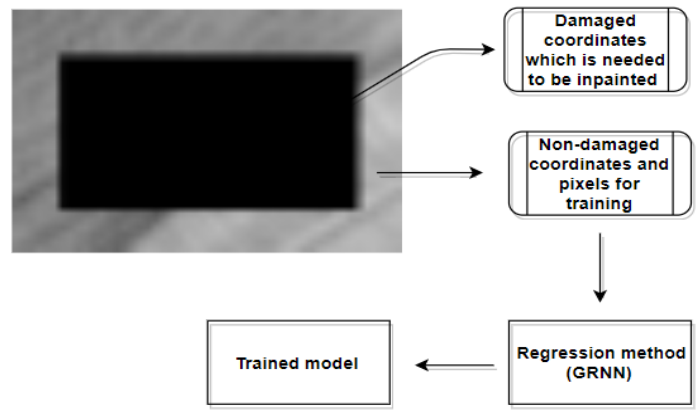

Figure.2

Algorithm 3:

Step1. Identifying the damaged parts from the damaged image (ref. Algorithm 2)

Step2. Feed the damaged part coordinates to the trained model one by one

Step3. Collect the output produced by the module with respect to input coordinates in an array.

Step4. Finally, Inpaint the damaged part of the Image using the outputs produced by the trained model.

## III. ALGORITHMS AND DATA BASE

### A) General Regression Neural Network(GRNN)[7]

The general regression neural network being the memory based neural network is designed to provide a measure of continuous variables and convergence to the original regression surface, it works on one pass learning principle. The ability to get trained of GRNN make it as a used an algorithm for image inpainting as in the case of image inpainting as the assumption of linearity is not assumed. GRNN provides major benefit with its fast learning capabilities and very less error with the increased availability of the training sample. For image inpainting the top-down model is followed by which missing regions are first identified and then sorted according to their sizes and then inpainting procedure is applied to all unknown pixel using the trained GRNN network. Since the image inpainting application doesn't need any learning which not the case of other application .In parallel, we can use some other neural network by overfitting and it makes the network highly sensitive. This overfitting and high sensitivity leads to memorization of learning hence GRNN helps in obtaining better results. The computational time achieved by GRNN is also less and is comparable to some of the most efficient approaches as Total Variation.

### B) Least Square-Support Vector Machine `[14]

A model is obtained to predict the damaged part which was done using the trained network; the network was trained using the data that has the strongest correlation with the damaged area. The LS_SVM model is improved using the additive function to make suitable and full use of correlate data in the image. The damages are made in accordance with the row, column and to cover both the row and columns and then the algorithm is compared to find the best fit for all the different kind of damages. The overall process was spliced into three key steps i.e. data selection, selecting kernel function for the support vector machine and then inpainting image using regression model.

*1) RD-Row direction*
In this algorithm we first identify all the damaged parts in a row and non-damaged part in the row is kept for training .So LS-SVM model is trained with all non-damaged part and using trained model we inpaint damaged pixels . Since the training phase is for row wise i.e., one-by-one row wise therefore one kernel is enough for obtaining better results.

*2) CD-column direction*
Same process of RD is carried out but in the column direction .Similar to RD ,CD also uses one kernal .

*3) RC-Row Column direction*
Row-Column is a process which includes both RD and CD method. Initially we inpaint the image using RD and CD, Then we average images obtained in both the method .Therefore the inpainting image will be influenced by both vertical and horizontal direction

*4) 2-kernal approch*
While using the basic LS-SVM model with 1-D kernal we only use the training data of single direction which can be either row(X) or column(Y) to train the model. By making the change in parameters of the kernal and using the data of both horizontal and vertical direction simultaneous lt to train the same kernal we can produce better results. This is done by simply making kernal function of x and y coordinates.So the model will havetwo input parameter those are  coordiantes (x,y) and output/target will be the corresonding pixel values .

### C) Image database:

This paper gives a comparative study of different algorithms which are implemented on 4 randomly selected image take from natural database i.e., Lena, baboon, House, and pepper are shown in figure. From the paper [14] the mask has been imitated and applied with four randomly selected images as shown in figure

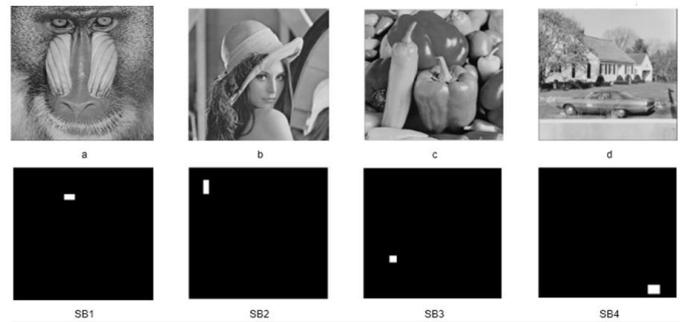

Figure.3

SB-Selective blocks
In Figure.3, a, b, c and d show the randomly selected natural images and SB1, SB2, SB3, SB4 are the mask which are used in LSSVM paper [14].

### D) Standard Metrics used

PSNR- Peak Signal to Noise Ratio.
$$\text{PSNR} = 10\log\frac{(2^n - 1)^2}{\text{MSE}} \quad (1)$$
Where MSE is mean Square Error, number of bits representing a pixel is denoted by n.

$$MSE = \frac{1}{M \times N}\sum_{i=1}^{N}\sum_{j=1}^{M}[I(i,j) - I'(i,j)]^2 \quad (2)$$
Where M and N are the size of the image, I(i, j) and
I(i ,j)' represent a pixels from original and inpainted image respectively.

## IV. RESULTS

### a) Damaged image VS inpainted image

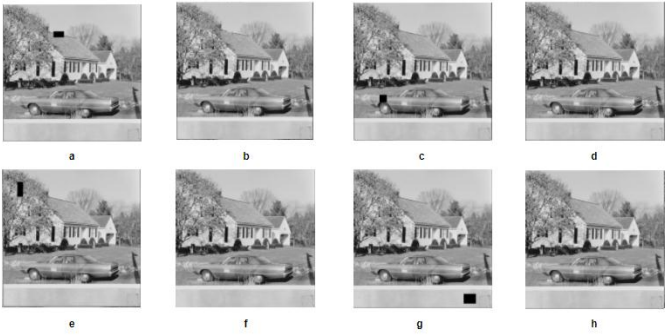

Figure.4

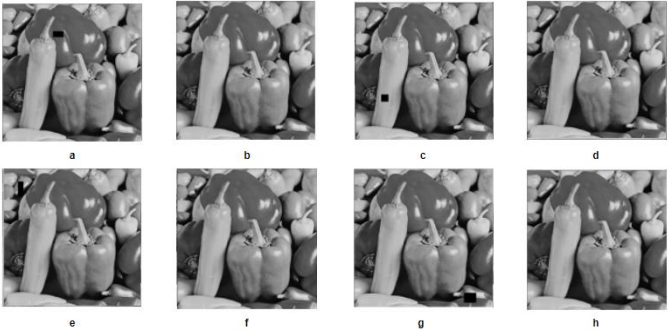

Figure.5

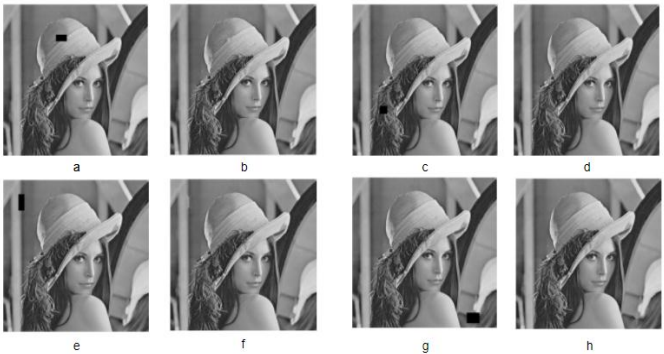

Figure.6

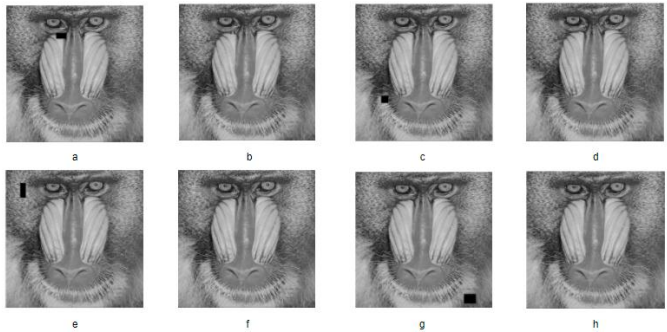

Figure.7

Figure.4, Figure.5, Figure.6 and Figure.7 shows the output of mask and inpainted images a-b, c-d, e-f, g-h in each figure determine masked and inpainted images respectively.

Figure.8, Figure.9, Figure.10 and Figure.11 shows the bar graph comparison of PSNR value obtained from original image vs inpainted image by varying radius which uses GRNN as the regression method. The comparison study of various radius has be limited up to 13 R because after that mere changes in PSNR values where noticed and in most cases it shows either no change or degraded PSNR values. SB indicates Selective block[14] and SB1,SB2,SB3 and SB4 are determined by 'blue', 'orange, 'Gray' and 'yellow' respectively.

### b) Psnr comparison of alogrithm

Table.1 Comparison Table

| Methods | RD [14] | CD[14] | RC[14] | LSSVM[14] | GRNN[7] |
|---|---|---|---|---|---|
| **Lena** | | | | | |
| SB | 44.1007 | 43.5723 | 43.2017 | **46.6916** | 39.0379 |
| SB1 | 41.2412 | **56.997** | 45.5484 | 45.0611 | 41.166 |
| SB2 | 42.8312 | 43.1817 | 42.5614 | **44.0992** | 42.2525 |
| SB3 | 34.8035 | 35.2013 | 35.4407 | 35.2389 | **36.3345** |
| **baboon** | | | | | |
| SB | 38.2699 | 39.4718 | 38.8752 | **40.6597** | 39.3112 |
| SB1 | 35.7969 | 37.6467 | 36.7042 | **40.2077** | 39.2543 |
| SB2 | **43.4070** | 39.5252 | 41.3444 | 42.1211 | 41.2241 |
| SB3 | 43.5079 | 44.5105 | 44.7857 | **45.5451** | 44.6120 |
| **House** | | | | | |
| SB | 43.7181 | 43.4011 | 43.6607 | **59.4127** | 44.0333 |
| SB1 | 38.3670 | 41.3158 | 39.7734 | **45.6924** | 40.3550 |
| SB2 | 48.4308 | 47.3452 | 46.4780 | 41.9367 | **46.0730** |
| SB3 | 55.7651 | 33.4928 | 39.1676 | 45.5703 | **59.1102** |
| **pepper** | | | | | |
| SB | 50.1607 | 47.4999 | 47.6796 | 58.4780 | **59.5475** |
| SB1 | 34.2458 | 33.7729 | 34.0507 | 33.0400 | **39.3019** |
| SB2 | 37.4710 | 61.0277 | 42.6293 | 62.0861 | **63.0271** |
| SB3 | 30.8244 | 29.2317 | 30.2859 | 40.1111 | **43.9614** |

Table.1 shows the comparison of two algorithm LSSVM and GRNN. LSSVM has four method, they are RD, CD, RC and 2 kernel. The maximum PSNR has been highlighted for easy identification.

### c) Psnr comparison of radial analysis in GRNN

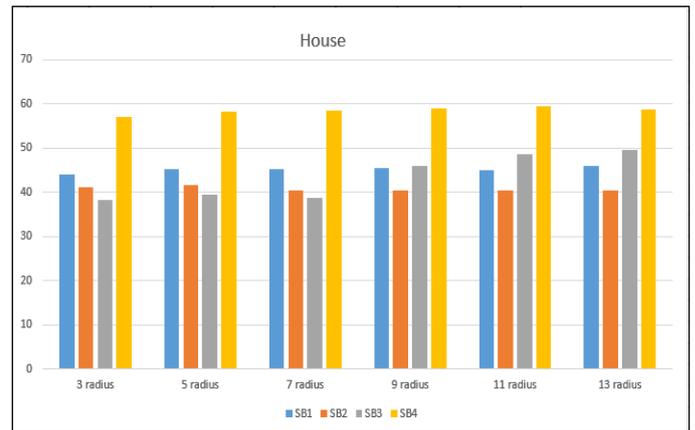

Figure.8

Figure.8, Figure.9, Figure.10 and Figure.11 shows the PSNR comparison of Radial base analysis applied on damaged images such as House, Pepper, Lena and Baboon respectively.

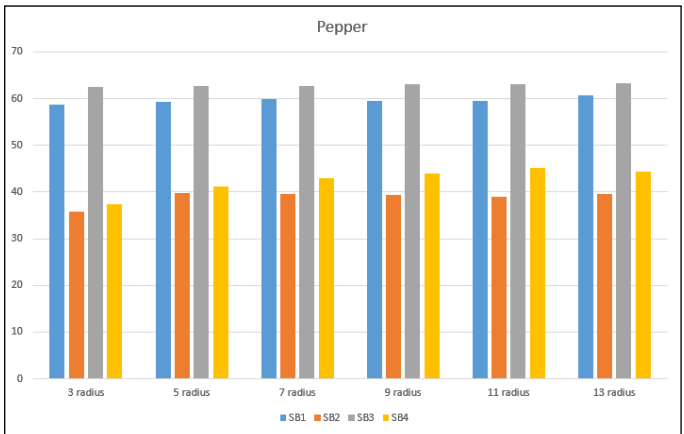

Figure.9

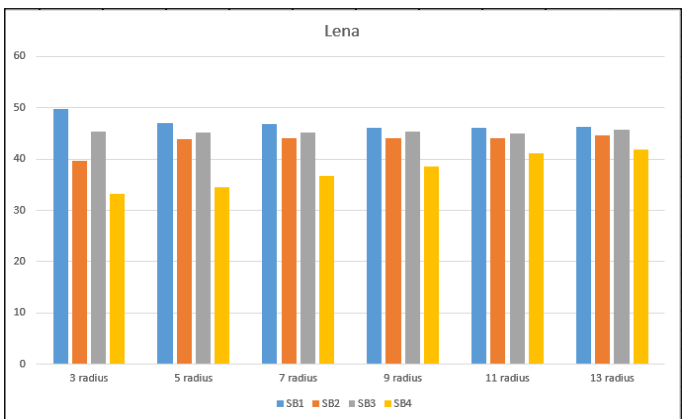

Figure.10

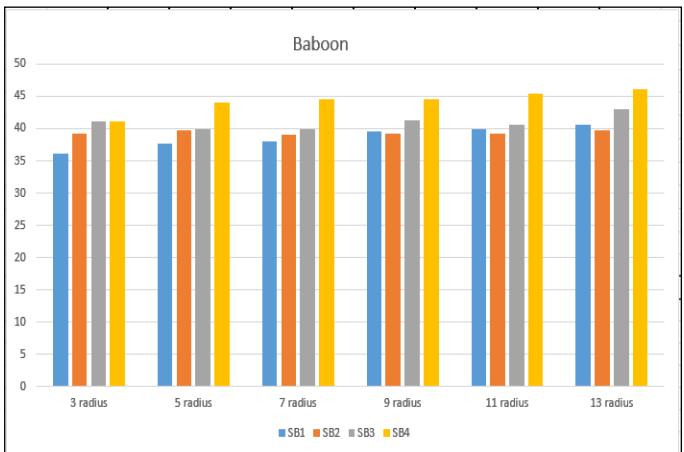

Figure.11

## V. CONCLUSION:

The paper successfully made an experimental based comparative analysis on two major regression algorithms they are LSSVM and GRNN in image inpainting and made a detail analysis on GRNN learning pixel selection using fast and feasible method called as radial analysis. Using this method we have huge verity of selecting learning pixels some are implemented and showed as bar graphs for easy understanding. According to the Table.1 GRNN shows the better results in Pepper image and LSSVM shows better in Baboon. In radial analysis .In SB1, PSNR value decreases as radius increases in Lena image and other images it either increases as radius increases or stays constant. In SB2, Lena and baboon images show increase or stay constant in PSNR values, House shows the maximum in 3 radius and minimum in 7 radius, pepper image shows minimum is 3 and maximum in 5.In SB3, Pepper and Lena images shows the no change and, House and Baboon PSNR value increases as radius increases. In SB4, all the image shows the progressive results as radius increases. This work can be further carried out with other regression algorithms such as FITNET, LSSVM etc.